\documentclass[sn-nature]{sn-jnl}

\usepackage{graphicx}%
\usepackage{multirow}%
\usepackage{amsmath,amssymb,amsfonts}%
\usepackage{amsthm}%
\usepackage{mathrsfs}%
\usepackage[title]{appendix}%
\usepackage{xcolor}%
\usepackage{textcomp}%
\usepackage{manyfoot}%
\usepackage{booktabs}%
\usepackage{algorithm}%
\usepackage{algorithmicx}%
\usepackage{algpseudocode}%
\usepackage{listings}%
\usepackage{makecell}

\theoremstyle{thmstyleone}%
\theoremstyle{thmstyletwo}%

\theoremstyle{thmstylethree}%

\begin{document}

\title[Article Title]{Estimating unknown parameters in differential equations with a reinforcement learning based PSO method}


\author[1,2]{\fnm{Wenkui} \sur{Sun}}\email{swk2022@mail.dlut.edu.cn}

\author[1]{\fnm{Xiaoya} \sur{Fan}}

\author[1]{\fnm{Lijuan} \sur{Jia}}

\author[3]{\fnm{Tinyi} \sur{Chu}}

\author[2,4]{\fnm{Shing-Tung} \sur{Yau}}

\author*[2,4]{\fnm{Rongling} \sur{Wu}}\email{ronglingwu@bimsa.cn}

\author*[1]{\fnm{Zhong} \sur{Wang}}\email{zhongwang@dlut.edu.cn}

\affil[1]{\orgdiv{School of Software Technology}, \orgname{Dalian University of Technology}, \orgaddress{\postcode{116024}, \state{Liaoning}, \country{China}}}

\affil[2]{\orgname{Beijing Institute of Mathematical Sciences and Applications}, \orgaddress{\postcode{101408}, \state{Beijing}, \country{China}}}

\affil[3]{\orgdiv{Meinig School of Biomedical Engineering}, \orgname{Cornell University}, \orgaddress{\postcode{NY 14853}, \state{Ithaca}, \country{United States}}}

\affil[4]{\orgdiv{Yau Mathematical Sciences Center}, \orgname{Tsinghua University}, \orgaddress{\postcode{100084}, \state{Beijing}, \country{China}}}



\abstract{Differential equations offer a foundational yet powerful framework for modeling interactions within complex dynamic systems and are widely applied across numerous scientific fields. One common challenge in this area is estimating the unknown parameters of these dynamic relationships. However, traditional numerical optimization methods rely on the selection of initial parameter values, making them prone to local optima. Meanwhile, deep learning and Bayesian methods require training models on specific differential equations, resulting in poor versatility. This paper reformulates the parameter estimation problem of differential equations as an optimization problem by introducing the concept of ``particles'' from the particle swarm optimization algorithm. In this framework, the solution is represented as a swarm of particles, each embodying a candidate solution through its position and velocity. The particles iteratively update through mutual interactions, facilitating convergence toward an optimal solution. Building on reinforcement learning-based particle swarm optimization (RLLPSO), this paper proposes a novel method, DERLPSO, for estimating unknown parameters of differential equations. We compared its performance on three typical ordinary differential equations with the state-of-the-art methods, including the RLLPSO algorithm, traditional numerical methods, deep learning approaches, and Bayesian methods. The experimental results demonstrate that our DERLPSO consistently outperforms other methods in terms of performance, achieving an average Mean Square Error of $1.13 \times 10^{-05}$, which reduces the error by approximately 4 orders of magnitude compared to other methods. 
Apart from ordinary differential equations, our DERLPSO also show great promise for estimating unknown parameters of partial differential equations. The DERLPSO method proposed in this paper has high accuracy, is independent of initial parameter values, and possesses strong versatility and stability. This work provides new insights into unknown parameter estimation for differential equations.}

\keywords{Differential Equations, Reinforcement Learning, Particle Swarm Optimization, Unknown Parameter Estimation}



\maketitle

\section{Introduction}\label{sec1}

At every moment, vast amounts of data are being collected through various human activities. Uncovering the hidden dynamics from these data is a fundamental yet challenging problem across many different fields \cite{bib1}.
Ordinary differential equations (ODEs) model the rates of change in dynamic processes across time or space. They are extensively used to describe complex systems in science, physics, economics, pharmacokinetics, neurophysiology, and systems biology \cite{bib2}.
Accurately estimating the parameters of equations is crucial in scientific research for drawing reliable and valid conclusions. Unknown or inaccurately estimated parameters can lead to results that misrepresent reality, hindering our understanding of scientific phenomena and laws. Therefore, accurately estimating these parameters before analyzing the system is particularly important to avoid such issues \cite{bib3}.
ODEs are prevalent in various research fields, but universally accepted methods for estimating their parameters are lacking. This limitation hampers our understanding and prediction of system behavior, highlighting the critical need to estimate unknown parameters in ODES.

There are two commonly used methods for parameter estimation in ODEs: numerical solution methods and non-parametric methods. Numerical methods use the least squares method to fit the ODE solutions to observed data, providing parameter estimates that accurately represent actual behavior. However, due to the lack of analytical solutions for many ODEs, these methods can be computationally intensive. In contrast, non-parametric methods bypass explicit ODE solutions, utilizing smoothing techniques for estimation. While this reduces computational overhead, it complicates optimization, making it more sensitive to noise and prone to converging on local optima. 
Edsberg et al. \cite{bib4} used numerical methods for parameter estimation, which have some drawbacks: using ODE solvers can increase computational complexity \cite{bib3}, and optimization methods may heavily rely on initial parameter values, making them prone to getting trapped in local optima.

To improve computational efficiency, Varah \cite{bib5}, Ramsay and Silverman \cite{bib6}, as well as Chen and Wu \cite{bib7}, proposed a simplified two-stage method to avoid the direct numerical solution of ODEs. In the first stage, the state functions and their first-order derivatives are estimated from observed data and treated as fixed variables. In the second stage, the parameters are estimated using standard least squares methods. This method performs well for simple ODEs. However, accuracy may diminish when estimating higher-order derivatives in many ODEs. To improve efficiency and increase estimation accuracy, Ramsay et al. \cite{bib8} proposed the parameter cascade method, a generalized smoothing technique. This method models the unknown solutions of ODEs as a linear combination of B-splines, applying penalized smoothing to the observed data. The penalty term, defined by the ODEs, helps prevent overfitting of the non-parametric function.

Recent advancements in Bayesian methods for parameter estimation in ODEs have been notable. Wakefield \cite{bib9} and Lunn \cite{bib10} applied Bayesian methods to pharmacokinetic models. However, the computational intensity of this Bayesian approach stems from the need to numerically solve the ODEs at each iteration of the Markov Chain Monte Carlo (MCMC) process, significantly affecting efficiency. Huang et al. \cite{bib11} proposed a Bayesian method that replaces ODEs constraints with probabilistic expressions and integrates them with non-parametric data fitting into a joint likelihood framework, using a MCMC sampling scheme. However, this method requires prior knowledge of the structure of ODEs, and may not yield satisfactory results for complex ODEs.

The development of machine learning algorithms has significantly enhanced the numerical solution and parameter estimation of ODEs. 
Brunton et al. \cite{bib12} combined sparse regression and machine learning with nonlinear dynamical systems to extract governing equations from noisy data, addressing the challenge of idenfitying control equations. 
However, this method relies on the chosen measurement variables and function basis. If chosen improperly, it may not be able to identify an accurate sparse model.
Raissi et al. \cite{bib13, bib14} used physics-informed neural networks (PINNs) to estimate parameters in various physical systems. In this approach, parameters are incorporated as part of the network training process, and once the network training is completed, the optimized parameter values can be obtained. Additionally, Neural Ordinary Differential Equations (Neural ODEs) \cite{bib15} extend deep neural networks to continuous-time applications, offering high memory efficiency, strong flexibility, and effectiveness in time series tasks. Challenges remain, however, in computational costs and parameter tuning, particularly for complex systems and large datasets, necessitating further optimization of model efficiency and training duration.

Arloff et al. \cite{bib16} proposed a two-step method to address that avoids numerical challenges in stiff ODEs via polynomial approximation and reduce the parameter search space using the Particle Swarm Optimization (PSO) algorithm. This approach is effective for complex stiff ODEs, identifying feasible solutions with reduced computational load. However, it only narrows the parameter space and the quality of its solution relies on the accuracy of the polynomial approximation. PSO is a swarm intelligence algorithm known for its strong global search capability, simple parameter configuration, rapid convergence, ease of implementation, robustness, and wide applicability in complex optimization problems. Wang et al. \cite{bib17} developed a large-scale optimization algorithm, RLLPSO, which integrates PSO with reinforcement learning (RL) to enhance the speed and accuracy of convergence.

We reformulates parameter estimation of differential equations as an optimization problem, and introduce a novel method, DERLPSO, to solve it. DERLSPO enhances RLLPSO with several novel strategies, i.e., logarithmic initialization, reinitialization mechanisms, and bottom-up update strategy, to achieve higher convegence speed, as well as more accurate and stable estimation. The DERLPSO method is independent of initial parameter values and exhibits high versatility and generalization capabilities. It has been tested on three types of ODEs: Lorenz, FitzHugh-Nagumo, and Lotka-Volterra equations, and three types of partial differential equations (PDEs): Heat, Transient convection-diffusion, and Helmholtz equations. The performance of DERLPSO has been compared with the state-of-the-art methods, including the RLLPSO algorithm, traditional numerical methods, deep learning approaches, and Bayesian methods.

The remainder of this paper is organized as follows. Section \ref{sec2} details the proposed algorithm. Section \ref{sec3} outlines the experimental procedures and discusses the results. Section \ref{sec4} concludes with insights for future research.

\section{DERLPSO Theory and Structure}\label{sec2}

Estimating unknown parameters in differential equations involves identifying undetermined constants or coefficients within the equations. The goal is to determine the specific values of these unknown parameters using available information, such as observational data, initial conditions, so that the differential equations accurately represent the dynamic behavior of the system. 
For example, in the Lotka-Volterra equations shown in Equation (\ref{eq1}), $x$ and $y$ are referred to as state variables, representing the populations of prey and predator, respectively. The derivatives of these variables, $\frac{dx}{dt}$ and $\frac{dy}{dt}$, denote the rates of change of these variables over time due to their interactions, with $t$ representing time. The parameters $\alpha$, $\beta$, $\delta$ and $\gamma$ are coefficients that characterize the interaction dynamics of the two species. This paper addresses the problem of determining the values of $\alpha$, $\beta$, $\delta$ and $\gamma$, given the values of $x$, $y$, and time $t$.

\subsection{DERLPSO Structure}\label{sub1sec2}

\begin{figure}[ht]
\centering
\includegraphics[width=0.9\textwidth]{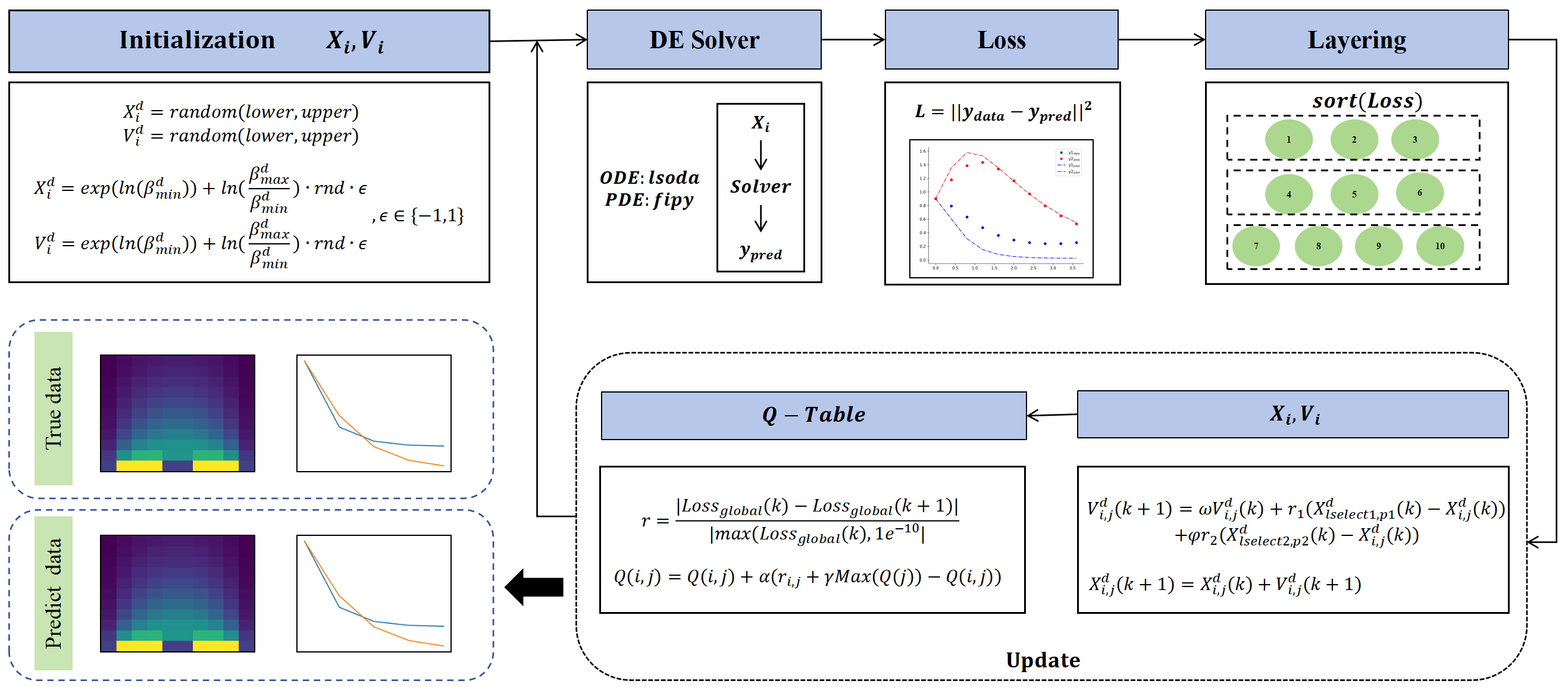}
\caption{Illustration of the proposed DERLPSO algorithm for estimating unknown parameters in differential equations. It is a reinforcement learning-based particle swam optimization approach, where each particle $i$ has a location $X_i$ representing a candidate estimation of unknown parameters, and a velocity $V_i$ determining its location for the next iteration. The particles are layered according to their losses, whose calculation rely on a differential equation solver tool, $lsoda$ for ODEs and $fipy$ for PDEs. A specific strategy is designed to update particles in different layers. The number of layers is determined by a reinforcement learning-based strategy. The Q-Table $Q$ is optimized to achieve the highest reward $r$.}\label{fig1}
\end{figure}

PSO is a swarm intelligence algorithm, known for its strong global search capability, simple parameter setup, fast convergence, ease of implementation, robustness, and broad applicability, making it a powerful tool for solving complex optimization problems. In this study, each particle represents a candidate parameter value for a differential equation, framing the parameter estimation problem as an optimization problem. Reinforcement learning is employed to enhance the global search capability of the PSO algorithm, thereby enabling PSO to converge to the optimal solution more quickly. RLLPSO \cite{bib17} is a large-scale optimization algorithm enhanced by reinforcement learning. Specifically, RLLPSO enhances population diversity by constructing a level-based swarm structure and utilizes a reinforcement learning strategy to dynamically control the number of population levels, thereby improving search efficiency. 
To further improve the convergence speed, stability, and accuracy of RLLPSO, this paper introduces logarithmic initialization, reinitialization mechanisms, and a bottom-up update strategy. A novel method, DERLPSO, is proposed for solving the unknown parameters in differential equations. Logarithmic initialization effectively covers the parameter space and enhance the sampling probability of small magnitude numbers, thereby improving the search efficiency of the algorithm. Reinitialization mechanisms provides new convergence opportunities by redefining particles' positions, facilitating quicker discovery of the global optimum. The bottom-up update strategy ensures that lower-level particles update based on the accurate states of higher-level particles, thereby avoiding error accumulation and further enhancing the convergence stability and accuracy of the algorithm.

The diagram of the DERLPSO algorithm is presented in Fig. \ref{fig1}. The location and velocity of each particle, $X_i$ and $V_i$ are first initialized. For the  Lotka-Volterra equations shown in Equation (\ref{eq1}), $X = [\alpha, \beta, \delta, \gamma]$ and $V = [\alpha_v, \beta_v, \delta_v, \gamma_v]$. Two random initialization strategies are used to ensure good coverage of the entire search space, which can improve the diversity of the solution. Then, the particles are updated iteratively. In each iteration, $LSODA$ or $FIPY$ is used solve the ODEs or PDEs given the candidate parameters corresponding to each particle. Mean Square Error(MSE) between the simulated data and the fitted data was calculated, according to which the particles are sorted and layered. The number of layers is determined by Q-learning algorithm. The layering operation allows diverse update of particles. Specifically, each particle was updated according to two particles randomly selected from the previous layers. The Q-table is updated afterwards. The iteration process stops until the predefined termination conditions are reached. The pseudocode of DERLPSO is presented in Algorithm \ref{alg1}.

\begin{algorithm}
      \renewcommand{\algorithmicrequire}{\textbf{Initialize:}}
      \caption{ DERLPSO }
      \label{alg1}
      \begin{algorithmic}[1]
        \Require Initialize the X and V of the particles swarm according to Eqs. (\ref{eq2}) and (\ref{eq3}), Q-Table
        \Require $ cur_{iterator} = 0$, $state_{current} = l_1$, $max_{iterator}$, threshold
        \State Calculate the Loss of each particles according to Eq. (\ref{eq10}) and update the global optimal particles gBest
        \While {termination conditions are not reached}
          \State Select the next action $acion_{next}$ according to Eq. (\ref{eq7})
          \State Sort the particles swarm based on Loss and divide it into $acion_{next}$ layers

          \For{$i=acion_{next}$; $i>=3$; $i--$}
            \For{$j=1$; $j<=sizeof(l_i)$; $j++$}
               \State Use Algorithm 2 to select two sample levels $l_{select1}, l_{select2}$
                \If{$l_{select1} == l_{select2}$}
                    \State Use Algorithm 2 to select two particles  $X_{lselect1,k1}, X_{lselect1,k2}$
                \EndIf
                \State Update the $X_{i,j}$ and $V_{i,j}$ of particles according to Eq. (\ref{eq5})
            \EndFor
          \EndFor
          \For{$j=1$; $j<=sizeof(l_1)$; $j++$}
                \State Use Algorithm 3 to select two particles $X_{l1,k1}, X_{l1,k2}$
                \State Update the $X_{2,j}$ and $V_{2,j}$ of particles according to Eq. (\ref{eq5})
          \EndFor
          \State Calculate the Loss of each particles according to Eq. (\ref{eq10}) and update global optimal particles gBest
          \State Calculate reward values r according to Eq. (\ref{eq8})
          \State Update Q-table according to Eq. (\ref{eq9})
          \If{$current_{iteration}$ == $max\_iterator$ / 2 and Loss(gBest) > threshold}
             \State Reinitialize the X and V of the particles swarm according to Eq. (\ref{eq3})
          \EndIf
          \State  Update $state_{action}$ = $action_{next}$, $current_{iteration}$ = $current_{iteration}$ + 1
        \EndWhile
      \end{algorithmic}
\end{algorithm}


\begin{equation}
    \begin{aligned}
        \frac{dx}{dt} &= \alpha x- \beta x \\
        \frac{dy}{dt} &= \delta x y- \gamma x
        \label{eq1}
    \end{aligned}
\end{equation}

\subsection{Initialization Strategy}\label{sub2sec2}
\begin{figure}[htbp]
\centering
\includegraphics[width=0.9\textwidth]{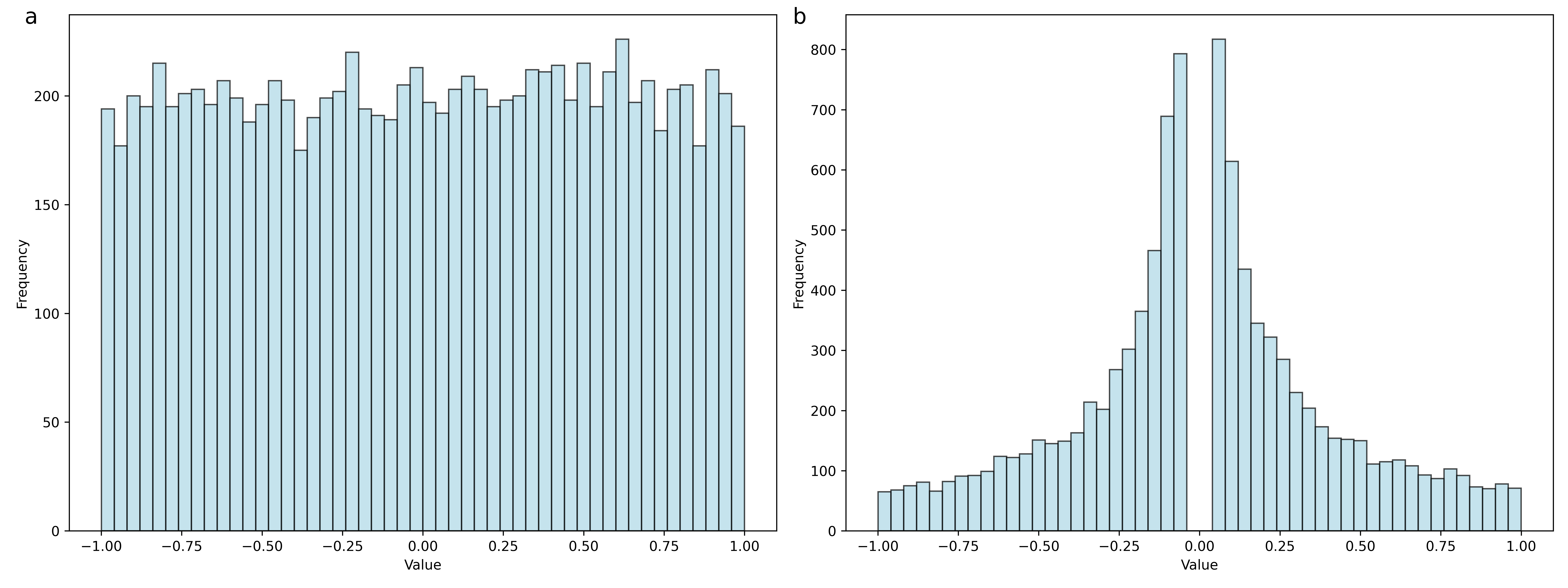}
\caption{Probability Distribution Plot. (a). Uniform Distribution with $\beta_{min} = -1$ and $\beta_{max} = 1$. (b). Logarithmic Distribution with $\beta_{min} = 0.05$ and $\beta_{max} = 1$}\label{fig2}
\end{figure}

This work employs two initialization strategies-\emph{Uniform} and \emph{logarithmic}, randomly sampling positions for half of the particles from each. For \emph{Uniform} strategy, the $d-th$ component is sampled with a uniform distribution in $[\beta_{min}^d, \beta_{max}^d$], as in Equation (\ref{eq2}), where $X_i^d$ represents the position of the $i^{th}$ particle in the $d^{th}$ dimension, $V_i^d$ represents the velocity of the $i^{th}$ particle in the $d^{th}$ dimension. Fig. \ref{fig2} a shows the probability distribution with $\beta_{min}^d=-1, \beta_{max}^d=1$.

\begin{equation}
    \begin{aligned}
        X_i^d=random(\beta_{min}^d, \beta_{max}^d) \\
        V_i^d=random(\beta_{min}^d, \beta_{max}^d)
        \label{eq2}
    \end{aligned}
\end{equation}

For \emph{Logarithmic} strategy \cite{bib18}, the $d^{th}$ component is sampled according to Equation (\ref{eq3}), where $rnd$ is a random number uniformly distributed in [0, 1), $\beta_{min}^d$ and $\beta_{max}^d$ represent the minimum and maximum values of the sampling range in the $d^{th}$ dimension.

\begin{equation}
    \begin{aligned}
        &X_{i}^{d} = \exp\left(\ln(\beta_{min}^d) + \ln\left(\frac{\beta_{max}^d}{\beta_{min}^d}\right) \cdot rnd\right) \cdot \epsilon \\
        &V_{i}^{d} = \exp\left(\ln(\beta_{min}^d) + \ln\left(\frac{\beta_{max}^d}{\beta_{min}^d}\right) \cdot rnd\right) \cdot \epsilon \\
    \end{aligned}
    \quad \text{where } \epsilon \in \{-1, 1\}
\label{eq3}
\end{equation}

Logarithmic sampling can provide a higher probability of sampling smaller magnitude values, and since logarithmic sampling can more effectively cover the entire parameter space, it can achieve the same coverage effect with fewer samples. However, since the logarithmic strategy cannot generate negative values, but the parameter values can be negative, to generate random numbers with both positive and negative values, a random factor of ±1 is multiplied by the random number generated by the logarithmic strategy. This is equivalent to introducing an additional sign randomness on the basis of the logarithmic strategy.

When $\beta_{min}=0.05$ and $\beta_{max}=1$, the logarithmic probability distribution is shown in Fig. \ref{fig2} b. It can be seen that there is a part of the values near zero that cannot be obtained. Therefore, it is important to set the  $\beta_{min}$ value as close to zero as possible, so as to reduce the range of values that cannot be randomly sampled.


\subsection{Particles Layering Strategy}\label{sub3sec2}

To maintain the diversity of candidate solutions in the particle swarm, the algorithm uses a level-based swarm structure. The particle swarm is divided into multiple levels, with higher levels representing particles closer to the optimal value. In this algorithm, a population of $N$ particles is divided into $L$ levels, based on the loss values of the candidate solutions, after being sorted in ascending order. The loss value represents the MSE between the simulated data and the fitted data for the current parameters. If the particles cannot be evenly divided, the remaining particles will be assigned to the last level, as shown in Equation (\ref{eq4}).

\begin{equation}
    \begin{aligned}
        LP_i=\begin{cases}\frac{N}{L},&i<L \\
        \frac{N}{L}+N\%L,&i=L\end{cases}
        \label{eq4}
    \end{aligned}
\end{equation}

\begin{figure}[htbp]
\centering
\includegraphics[width=0.7\textwidth]{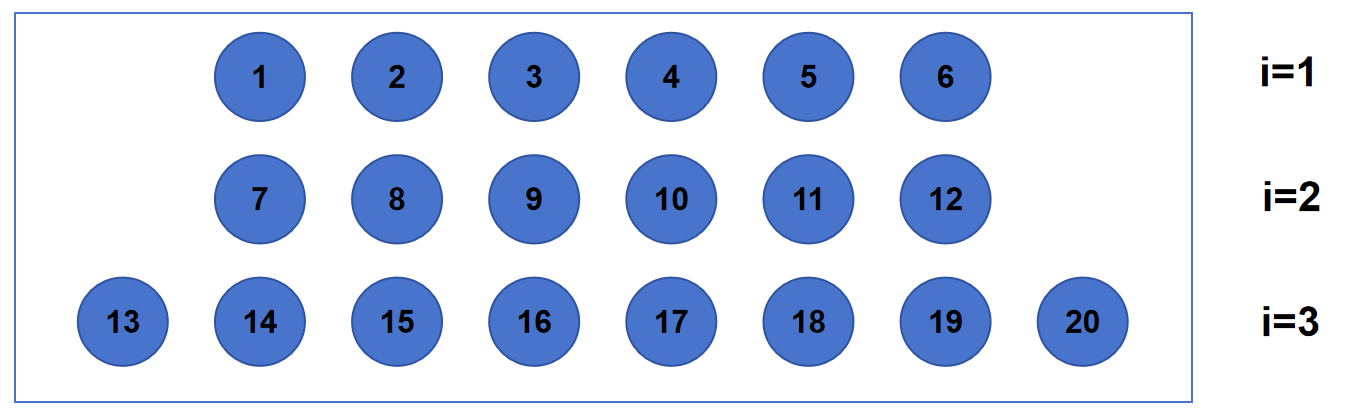}
\caption{Level-based structure of the particle swarm when $N = 20$ and $L = 3$.}\label{fig4}
\end{figure}

For example, when there are 20 particles divided into three levels, the first and second levels will have an equal number of particles, which is 6, and the third level will have 8 particles, as shown in Fig. \ref{fig4}.

During the iterative process, particles in the first level will not be updated, particles in the second level will randomly select two particles from the first level as samples for updating, and particles from other levels will randomly select two particles as samples from levels that are higher than the current level during the iteration. After selecting the samples, particles will be updated according to Equation (\ref{eq5}),

\begin{equation}
    \begin{aligned}
        &V_{i,j}^d(k+1)=\omega V_{i,j}^d(k)+r_1\left(X_{l\text{select}1,p1}^d(k)-X_{i,j}^d(k)\right)+\varphi r_2\big(X_{l\text{select}2,p2}^d(k)-X_{i,j}^d(k)\big) \\
        &X_{i,j}^d(k+1)=X_{i,j}^d(k)+V_{i,j}^d(k+1)
        \end{aligned}
        \label{eq5}
\end{equation}

where $X_{i,j}^d(k)$ represents the value of the j-th particle in the i-th level in the d-th dimension at the k-th iteration, and $V_{i,j}^d(k)$ represents its velocity. $X_{l_{select1}, p_1}$ and $X_{l_{select2}, p_2}$ represent particles randomly selected from levels $l_{select1}$ and $l_{select2}$, respectively, and $l_{select1} < l_{select2} < i$. For the $\omega $, a common setting is to linearly decrease it from 0.9 to 0.4 as the number of iterations increases \cite{bib19}. However, through experiments, it was found that setting $\omega $ as a uniformly distributed random variable within the interval [0, 1] can significantly improve search efficiency. Therefore, this paper adopts this method for selecting the $\omega$. Parameters $r_1$ and $r_2$ are uniformly distributed random values within the interval [0, 1]. Through the randomness of  $r_1$ and $r_2$, the particle's motion trajectory has greater randomness and diversity, which allows the particle swarm to more comprehensively explore the solution space and increases the chances of finding the true optimal solution during the search process. The parameter $\phi$ controls the influence of the sample with the relatively higher loss value between the two samples on the current particles, and its value should be within the interval (0, 1]. 

Since lower-level particles (with bigger level numbers) rely on higher-level particles (with smaller level numbers) for updates, if higher-level particles update and do not decrease their loss values, but the lower-level particles still rely on them for their updates, this leads to the lower-level particles being updated in the wrong direction. Therefore, in the process of updating particles, a bottom-up method is used, where particles are updated starting from the lowest levels.

\subsection{Tier Selection Strategy}\label{sub4sec2}

By employing level-based swarm structure, the particle swarm can maintain a good diversity, with lower-level particles having more potential learning samples, and higher-level focusing more on exploring the optimal value. However, randomly selecting two particles as samples will impose certain limitations on the convergence efficiency of the algorithm, as particles update themselves by moving closer to the particle with a lower loss value with a greater weight. To improve convergence efficiency, a competitive mechanism is introduced to increase the probability of selecting particles with lower loss values as samples for updates, and the probability of triggering this mechanism increases with the number of iterations, as shown in Equation (\ref{eq6}),

\begin{equation}
p=(\frac{cur_{iterator}}{max_{iterator}})^2
\label{eq6}
\end{equation}

where $p$ represents the probability of triggering the level competition mechanism during iteration, $cur_{iterator}$ represents the current iteration number, and $max_{iterator}$ represents the maximum iteration number. First, calculate the threshold $p$. In the subsequent two loops, a random number $rand$ is generated each time. If the random number is less than $p$, trigger the following mechanism: randomly select two levels that are higher than the current particle's level, and choose the higher one. If not, a level is randomly selected that is higher than the current particle's level. After the two loops, two levels are obtained. Finally, a particle is randomly selected from each of the two levels as a sample. 
This increases the probability of selecting particles with lower loss values, which can accelerate the convergence of the algorithm, as shown in Algorithm \ref{alg2}.

\begin{algorithm}
    \renewcommand{\algorithmicrequire}{\textbf{Input:}}
    \renewcommand{\algorithmicensure}{\textbf{Output:}}
    \caption{ Layer selection }
    \label{alg2}
    \begin{algorithmic}[1]
    \Require
     Current level $l_{current}$
    \Ensure Two sample levels
    \State Calculate p according to Eq. (\ref{eq6}) to determine if the competition mechanism is triggered
    \For{$k=1$; $k<=2$; $k++$}
        \If{$rand < p$}
            \State Randomly select two levels $l_{r1}$,$l_{r2}$ higher than $l_{current}$
            \State Choose the higher of $l_{r1}$ and $l_{r2}$ as $l_{selectk}$
        \Else
            \State Randomly select one level $l_{selectk}$ higher than $l_{current}$
        \EndIf
     \EndFor
    \If{$l_{select1} < l_{select2}$}
        \State Swap $l_{select1}$ and $l_{select2}$
    \EndIf \\
    \Return $l_{select1},l_{select2}$
    \end{algorithmic}
\end{algorithm}

However, this mechanism can not guarantee that the two selected levels are different. When the two selected levels are the same, the two randomly selected sample particles may have the situation where the loss value of $X_{l\text{select1},p1}$ is greater than that of $X_{l\text{select2},p2}$. According to Equation (\ref{eq5}), it can be observed that the sample $X_{l\text{select1},p1}$ has a greater influence on the current particle than sample $X_{l\text{select2},p2}$, which means that the current particle will not move in the direction of a better particle. Therefore, when the two selected levels are the same, further processing is needed. In this case, a particle is first randomly selected from the given level and when selecting the second particle, it is randomly selected from the particles that come after the previously selected particle. This ensures that the loss value of $X_{l\text{select1},p1}$ is less than $X_{l\text{select2},p2}$, as shown in Algorithm \ref{alg3}.

\begin{algorithm}
     \renewcommand{\algorithmicrequire}{\textbf{Input:}}
     \renewcommand{\algorithmicensure}{\textbf{Output:}}
      \caption{ Particles selection }
      \label{alg3}
      \begin{algorithmic}[1]
        \Require
        Choose the of particles from level $l_{select}$
        \Ensure Two sample particles
        \State index1 = $l_{select}$[range from 0 to (sizeof($l_{select}$) - 2)]
        \State index2 = $l_{select}$[range from (index1 + 1) to (sizeof($l_{select}$) - 1)]
        \State $X_{lselect,p1} = X[index1]$
        \State $X_{lselect,p2} = X[index2]$ \\
        \Return $X_{lselect,p1},X_{lselect,p2}$
      \end{algorithmic}
    \end{algorithm}

\subsection{Reinforcement Learning-guided Particle Layering}\label{sub5sec2}

Reinforcement Learning is a type of machine learning method where an intelligent agent learns to make decisions by interacting with the environment, with the goal of maximizing cumulative rewards. Unlike supervised learning, reinforcement learning does not rely on explicit labeled data, but instead learns through trial and error by receiving feedback. In each step, the intelligent agent selects an action based on the current state, and the environment responds with a reward or penalty. The agent adjusts its strategy through this feedback, optimizing its decision-making process. Q-learning is a widely used reinforcement learning algorithm. 

Q-learning is a value function-based reinforcement learning algorithm, and Q-table is a lookup table in Q-learning that stores the Q-values corresponding to each state-action pair, in the form of a table or matrix. The Q-values represents the expected cumulative rewards that an agent will receive in the future when taking a certain action in a specific state. The core function of the Q-table is to help the agent select the optimal action based on its current state. 
The agent selects an action according to the Q-table in each state, and updates the Q-values based on the reward received and the next state. Through repeated iterations, the agent gradually learns to select the optimal policy that maximizes the cumulative reward. The key to Q-learning is to balance exploration and exploitation using an $\epsilon$-greedy strategy, and to optimize the Q-values step by step using the Bellman equation until convergence. As shown in Algorithm \ref{alg4}.

\begin{algorithm}
     \renewcommand{\algorithmicrequire}{\textbf{Initialize:}}
      \caption{ Q-Learning }
      \label{alg4}
      \begin{algorithmic}[1]
        \Require
        Initialize Q-Table, initial state s
        \For{each episode}
        \State With probability $\varepsilon$, choose a random action $a$
        \State Otherwise, choose $a = \arg\max Q(s, a)$
        \State Take action $a$, observe reward $r$ and next state $s'$
        \State Update Q-value:\[
        Q(s, a) = Q(s, a) + \alpha \left[ r + \gamma \max_{a'} Q(s', a') - Q(s, a) \right]
        \]
        \State Set $s = s'$
        \EndFor
      \end{algorithmic}
    \end{algorithm}

\begin{figure}[htbp]
\centering
\includegraphics[width=0.5\textwidth]{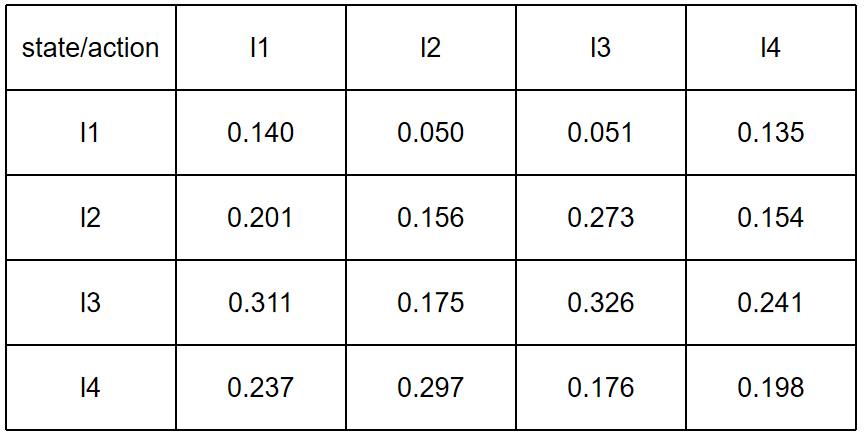}
\caption{ Q-table after 50 iterations with 4 layers.}\label{fig5}
\end{figure}

DERLPSO adapts to select different numbers of levels during the iteration process. This behavior is guided by the completion of the Q-learning algorithm of reinforcement learning. The core of the Q-learning algorithm is the Q-table, which is an $n$*$n$ matrix, where $n$ represents the number of candidate levels. Fig. \ref{fig5} shows the Q-table after 50 iterations with 4 candidate level numbers. In each iteration, the corresponding level number is selected based on the Q-table. After selecting the level number, each particle is reassigned to its respective level, and then each particle is updated. 
The values from l1 to l4 represent the candidate level numbers, indicating how many levels the particle swarm should be divided into, which can be 2, 4, 6, or 8, and so on. Of course, more candidate level numbers can be set. The initial Q-value for each state and action is set to 0. During action selection, a level number is chosen either randomly with a certain probability or by selecting the candidate level number with the highest Q-value, as shown in Equation (\ref{eq7}).

\begin{equation}
    action_{next}=
    \begin{cases}
        argmax(Q[state_{current},action_{current}]), &rand() < \epsilon \\
        action_{random}, &otherwise
    \end{cases}
    \label{eq7}
\end{equation}

In this algorithm, the reward value is not set to a fixed value, but rather calculated based on the loss value of the global optimal particle, as shown in Equation (\ref{eq8}),
\begin{equation}
    r=\frac{|Loss_{global}(k)-Loss_{global}(k-1)|}{|max(Loss_{global}(k),1e^{-10})|}
    \label{eq8}
\end{equation}

where, $Loss_{global}(k)$ and $Loss_{global}(k-1)$ represent the loss values of the global optimal solution at the current and previous iterations, respectively. During calculation, it is possible for the loss value of the optimal particle to be 0, so to avoid calculation errors, the larger value between $Loss_{global}(k)$ and $1e^{-10}$ is selected as the numerator when calculating the reward value. 

After receiving the reward, to maintain the effectiveness of the action selection strategy, the reinforcement learning policy updates Q-value through Equation (\ref{eq9}),

\begin{equation}
    Q(i,j)=Q(i,j)+\alpha(r_{i,j}+\gamma Max(Q(j))-Q(i,j))
    \label{eq9}
\end{equation}

where, $Q(i, j)$ represents the Q-value of transitioning from level $i$ to level $j$, and $\text{Max}(Q(j))$ represents the maximum Q-value of the available actions when the level is $j$. $\alpha$ is the learning rate, which controls the degree of influence of the new behavior on the original Q-value, and $\gamma$ is the discount factor, which weights the relative importance of immediate and future expected rewards.

\subsection{Reinitialize Strategy}\label{sub6sec2}

Tian et al. proposed that reinitializing some poorly performing particles during iteration can lead to better convergence \cite{bib20}. In the method presented in this paper, if the loss value of the global optimal particle does not reach the set threshold by the halfway point of the maximum iteration, all particles will be reinitialized based on the logarithmic strategy. Since the initialization of particles is a stochastic process, there is a possibility that particles may get trapped in local optima. Reinitializing the particles can help the algorithm escape local optima, explore a broader search space and increasing the likelihood of finding the global optimal solution.

\subsection{Particle evaluation}\label{sub7sec2}

To evaluate the performance of the unknown parameters represented by each particle, the LSODA \cite{bib21} method or the FiPy library is used to simulate the solution vector of the differential equations under the current predicted parameters, the values of $x$ and $y$ in Equation (\ref{eq1}), and compare it with the true solution vector. MSE is used to evaluate the quality of the current predicted parameters (the position information of the current particle), as shown in Equation (\ref{eq10}),

\begin{equation}
    MSE= \frac{1}{n} \sum_{i=1}^n(y_{true}-y_{pred})^2
    \label{eq10}
\end{equation}

where, $y_{true}$ represents the true solution vector of the differential equations, $y_{pred}$ represents the solution vector of the differential equations simulated with the current parameters, and $n$ denotes the length of the solution vector. A smaller MSE value indicates that the current predicted parameters are closer to the true parameters.

\section{Experiments and Results}\label{sec3}

\subsection{Differential Equations}\label{sub1sec3}

This paper validates the validates the proposed method using three common ODEs and three common PDEs. The six equations are: Lotka-Volterra equation, Lorenz equation, FitzHugh-Nagumo equation, Heat equation, Transient convection-diffusion equation, the Helmholtz equation.

\subsubsection{Lotka-Volterra Equation}\label{sub1sub1sec3}

The Lotka-Volterra equation originates from a classic biological system model\cite{bib22}, which is often used to describe the dynamic interactions between predators and prey in biological systems, specifically the fluctuations in the population sizes of both species. These equations were independently proposed by Lotka and Volterra in 1925 and 1926, respectively. The specific structure is shown in Equation (\ref{eq1}), where, $y$ represents the number of predators, $x$ represents the number of prey, and $\ {dy}{dt}$ and $\frac{dx}{dt}$ denote the rates of change of predator and prey populations, respectively. $t$ represents time, while $\alpha$, $\beta$, $\delta$, and $\gamma$ are parameters related to the interaction between the two species, all of which are positive real numbers. In this experiment, $\alpha$, $\beta$, $\delta$, and $\gamma$ are the unknown parameters. 

The variation between the two variables in this ODEs exhibits periodicity, aligning with the natural developmental patterns observed in biological systems. As the prey population increases, the predator population also grows, as the prey provides an abundant food source for the predators. However, when the predator population reaches a certain threshold, the prey population begins to decline because the predation rate exceeds the prey's reproduction rate. As the prey population decreases, predators experience a shortage of food, leading to a gradual decline in their numbers. This reduction in predator pressure allows the prey population to recover. Subsequently, the predator population increases again as the prey population rises. This cyclical dynamic repeats indefinitely.

\subsubsection{Lorenz Equation}\label{sub2sub1sec3}

The Lorenz equation are a simplified set of differential equations that describe the motion of fluid convection in the atmosphere \cite{bib23}. The Lorenz equation were proposed by the American meteorologist Edward Lorenz. He applied a Fourier expansion to a set of nonlinear ODEs that describe atmospheric thermal convection. By truncating the expansion, he derived a three-dimensional autonomous dynamical system, known as the Lorenz equation, which represent the expansion coefficients for vertical velocity, temperature difference between the top and bottom layers, denoted as $x$, $y$, and $z$, respectively. This equation is used to model the behavior of atmospheric convection. 

The specific structure of the equation can be seen in Equation (\ref{eq11}). These equations describe the rate of change of three quantities over time: $x$, which is proportional to the convective rate, $y$, which is proportional to the horizontal temperature variation, and $z$, which is proportional to the vertical temperature variation. The constants $\sigma$, $r$, and $\beta$ are system parameters that are proportional to the Prandtl number, Rayleigh number, and certain physical dimensions of the layer. In this experiment, $\sigma$, $r$, and $\beta$ are unknown parameters.

\begin{equation}
    \begin{cases}
        \frac{dx}{dt}=\sigma(y-x) \\
        \frac{dy}{dt}=rx-xz-y \\
        \frac{dz}{dt}=-\beta z+xy
    \end{cases}
    \label{eq11}
\end{equation}

\subsubsection{FitzHugh-Nagumo Equation}\label{sub3sub1sec3}

The FitzHugh-Nagumo equation are predominantly utilized to model the dynamics of spiking neurons \cite{bib24}, with the detailed structure of the equation presented in Equation (\ref{eq12}). In this experiment, $\theta_0$ and $\theta_1$ are unknown parameters.

\begin{equation}
    \begin{cases}
        \frac{du}{dt}=\gamma(u-\frac{u^3}3+v+\xi \\
        \frac{dv}{dt}=-\frac1\gamma(u-\theta_0+\theta_1v)
    \end{cases}
    \label{eq12}
\end{equation}

\subsubsection{Heat Equation}\label{sub4sub1sec3}

The Heat equation is a PDEs that describes the process of heat transfer. It is also known as the Fourier equation, named after the French mathematician and physicist Joseph Fourier. This equation has broad applications in physics, engineering, and many other fields, such as temperature distribution, heat transfer, and thermal conduction in materials. The law asserts that, in a homogeneous medium, the rate of heat transfer is proportional to the temperature gradient and occurs in the direction opposite to that gradient. The mathematical form of the heat conduction equation is typically represented as shown in Equation (\ref{eq13}).

\begin{equation}
    \frac{\partial u}{\partial t}=\alpha\nabla^2u
    \label{eq13}
\end{equation}

In this paper, a one-dimensional heat conduction equation is used, where $u(x, t)$ represents the temperature at position $x$ and time $t$, and $\frac{\partial u}{\partial t}$ denotes the rate of change of temperature with respect to time $t$. The parameter $\alpha$ represents the thermal diffusivity, a material property that describes the rate at which heat diffuses through the medium. The symbol $\nabla^2$  represents the Laplace operator, which in one-dimensional space is expressed as shown in Equation (\ref{eq14}). In this experiment, $\alpha$ is the unknown parameter.

\begin{equation}
    \nabla^2u=\frac{\partial^2u}{\partial x^2}
    \label{eq14}
\end{equation}

\subsubsection{Transient Convection-Diffusion Equation}\label{sub5sub1sec3}

The Transient Convection-Diffusion equation describes the temporal variation of a physical quantity in one-dimensional space under the combined influence of convection and diffusion. This equation has wide applications in fields such as fluid mechanics, heat conduction, and environmental science. The equation used in this paper is presented in Equation (\ref{eq15}),

\begin{equation}
    \frac{\partial u}{\partial t}+v\frac{\partial u}{\partial x}=\mathrm{D}\frac{\partial^2u}{\partial x^2}
    \label{eq15}
\end{equation}

where, $u(x, t)$ is the physical quantity to be solved, which is a function of spatial position $x$ and time $t$. The term $\frac{\partial u}{\partial t}$ represents the transient term, indicating the rate of change of the physical quantity over time. The term $v \frac{\partial u}{\partial x}$ is the convection term, where $v$ is the convection coefficient, describing the influence of convection on the spatial transport of $u$. The term $D \frac{\partial^2 u}{\partial x^2}$ is the diffusion term, $D$ is the diffusion coefficient, describing the effect of diffusion on the spatial distribution of $u$. In this experiment, $v$ and $D$ are the unknown parameters.

\subsubsection{Helmholtz Equation}\label{sub6sub1sec3}

The Helmholtz equation is named after the German physicist Hermann von Helmholtz. It describes wave phenomena, specifically the spatial distribution of wave fields (such as electromagnetic or acoustic fields) when wave propagation is subject to some form of linear constraint or restriction. These constraints can arise from boundary conditions (such as reflection and interference of waves in a finite region) or from inhomogeneities in the medium (such as the propagation of sound waves in a layered medium). The Helmholtz equation is given in Equation (\ref{eq16}),

\begin{equation}
    \frac{\partial^2u}{\partial x^2}+\frac{\partial^2u}{\partial y^2}+k^2u=0
    \label{eq16}
\end{equation}

where, $\phi$ represents the physical quantity to be solved (such as the potential function in an electromagnetic field or the pressure field in a sound wave), which describes the spatial distribution of the field. It is a function of $x$ and $y$. The constant $k$ is typically related to the physical context of the problem. In this paper, the value of $k$ is determined according to Equation (\ref{eq17}),

\begin{equation}
    k = \frac{2 \pi}{\lambda}
    \label{eq17}
\end{equation}

where, $\lambda$ represents the wavelength. In this experiment, $\lambda$ is the unknown parameter.

\subsection{Model Parameters}\label{sub2sec3}

The experimental data used was not generated from a fixed differential equations, but rather by randomly selecting unknown parameters from a certain distribution, in order to increase the randomness and generality of the data.

Considering that different ODEs have different data scales, the time range and initial points for simulating data were set differently. In the experiment, to verify the model's prediction performance on time series data of different lengths, experiments were conducted on data with 5, 8, and 10 time points, respectively, with specific parameters are shown in Table \ref{tab1}.


\begin{table}[htbp]
\caption{Parameters related to simulation data of three types of ODEs}\label{tab1}%
\begin{tabular}{@{}cccc@{}}
\toprule
ODEs & Lotka-Volterra  & Lorenz & FitzHugh-Nagumo\\
\midrule
Range of t    & [0,4]   & [0,4]  & [0,20]  \\
Points of t    & \multicolumn{3}{c}{5,8,10}  \\
Initial Point    & [0.9,0.9]   & [0,1,1.25]  & [0d,0]  \\
\makecell{Distribution of \\ Unknown Parameters}   & \makecell{$\alpha\sim N(0.4,0.5)$ \\ $\beta\sim N(1.3,0.5)$ \\ $\delta\sim N(1,0.5)$ \\ $\gamma\sim N(1,0.5)$}    
                                                      & \makecell{$\sigma\sim N(2,0.5)$ \\ $\beta\sim N(4,0.5)$ \\ $r\sim N(1,0.5)$}                           
                                                      & \makecell{$\theta_0\sim N(0.7,0.5)$ \\ $\theta_1\sim N(0.8,0.5)$}  \\           
\botrule
\end{tabular}
\end{table}

For PDEs, experiments were conducted separately for data with different spatial and temporal ranges to validate the model's prediction effectiveness on various data. The unknown parameters of the equations were set using truncated normal distributions, constrained within the interval (0,1]. The specific parameters are shown in Table \ref{tab2}.

\begin{table}[htbp]
\caption{Parameters related to simulation data of three types of PDEs.}\label{tab2}%
\begin{tabular}{@{}cccc@{}}
\toprule
PDEs & Heat  & Transient Convection-Diffusion & Helmholtz\\
\midrule
Range of t     & \multicolumn{2}{c}{[0,1]}     & /  \\
Points of t    & \multicolumn{2}{c}{10,20,40}  & /  \\
Range of x     & \multicolumn{3}{c}{[0,1]}         \\
Points of x    & \multicolumn{3}{c}{10,20,40}       \\
Range of y     & \multicolumn{2}{c}{/}         & [0,1]  \\
Points of y    & \multicolumn{2}{c}{/}         & 10,20,40  \\
\makecell{Distribution of \\ Unknown Parameters}   & \makecell{$\alpha\sim N(0.4,0.5)$}    
                                                   & \makecell{$D \sim N(0.5,0.5)$ \\ $v \sim N(0.5,0.5)$}
                                                   & \makecell{$\lambda\sim N(0.5,0.5)$}  \\           
\botrule
\end{tabular}
\end{table}

In the experiments, the population size $N$ for the algorithm was set to 100. In Equation (\ref{eq2}), the $lower$ was set to -10 and the $upper$ to 10. In Equation (\ref{eq3}), the $\beta_{min}$ was set to $1 \times 10^{-10}$ and the $\beta_{max}$ to 10, ensuring that the initial parameter values of the particles are constrained within the range of [-10, 10]. In Equation (\ref{eq5}), the parameter $\phi$ is assigned a value of 0.4, and in Equation (\ref{eq7}), $\epsilon$ is set to 0.9. For Equation (\ref{eq9}), the parameters $\alpha$ and $\gamma$ are defined as 0.4 and 0.8, respectively. The list of candidate levels is given by $Level = {4, 6, 8, 10}$. The reinitialization threshold is set to $1 \times 10^{-4}$. The maximum number of iterations is adjusted according to the specific equations, as shown in Table \ref{tab3}.

\begin{table}[htbp]
\caption{Parameters related to DERLPSO.}\label{tab3}%
\begin{tabular}{@{}cc@{}}
\toprule
Parameter  & Parameter Value\\
\midrule
lower             &     -10    \\
upper             &     10    \\
$\beta_{min}$     &     $1 \times 10^{-10}$    \\
$\beta_{max}$     &     10    \\
$\phi$            &     0.4    \\
$\epsilon$        &     0.9    \\
$\alpha$          &     0.4    \\
$\gamma$          &     0.8    \\
Level             &     \{4,6,8,10\}    \\
threshold         &     $1 \times 10^{-4}$    \\
\multirow{6}{*}{max\_iterator}  & Lorenz: 100          \\ 
                                & FitzHugh-Nagumo: 100 \\ 
                                & Lotka-Volterra: 200  \\
                                & Heat: 50          \\ 
                                & Transient Convection-Diffusion: 50 \\ 
                                & Helmholtz: 50  \\
\botrule
\end{tabular}
\end{table}

To validate the effectiveness of the proposed method for solving differential equation parameters, this paper compared the results of RLLPSO, numerical methods, deep learning methods, and Bayesian methods with the proposed DERLPSO method.

\subsection{DERLPSO for ODEs examples}\label{sub3sec3}

When evaluating model performance, this paper adopts two important indicators: MSE and Standard Deviation (SD). MSE emphasizes the average squared difference between predicted and true values, giving a larger penalty to larger errors, and can reflect the overall prediction bias of the model. SD is mainly used to measure the dispersion or width of a set of data, reflecting how spread out the data points are from the mean value. 
For the ODEs parameter estimation problem in this paper, SD is calculated for the errors between the true parameters and the predicted parameters. A smaller SD indicates that the differences are relatively clustered, reflecting that the prediction results are stable and consistent. By comprehensively analyzing these two indicators, we can thoroughly and deeply evaluate the performance of the constructed model, gaining a more complete understanding of its strengths and weaknesses.

\begin{figure}[htbp]
\centering
\includegraphics[width=0.9\textwidth]{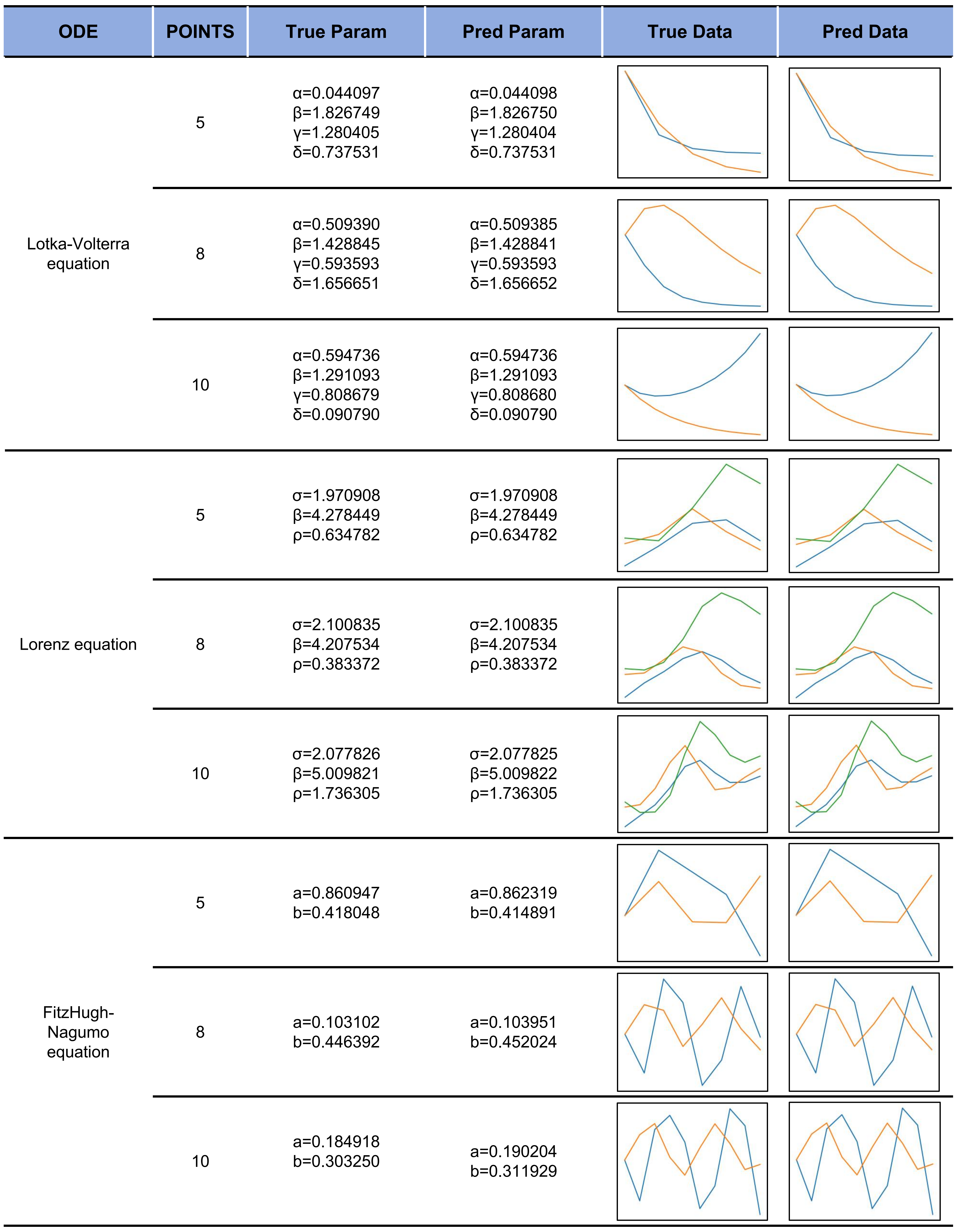}
\caption{Comparison of True and Predicted Data in ODEs Examples and Corresponding Experimental Scenarios.}\label{fig6}
\end{figure}

Fig. \ref{fig6} shows the comparison between the simulated data and the fitted data for different ODEs in corresponding experimental scenarios. Tables \ref{tab4}, \ref{tab5}, and \ref{tab6} show the mean of MSE across 100 sets of predicted and true parameters for the Lotka-Volterra equation, Lorenz equation, and FitzHugh-Nagumo equation. The values in parentheses represent the SD of the MSE across these 100 sets. The MSE reported here is the average MSE of multiple parameters. It can be seen that DEPLPSO provides reasonable predictions for the unknown parameters of the three types of ODEs. Both MSE and SD being relatively small. This indicates that the method has high accuracy and stability.

\begin{table}[htbp]
\caption{Summarize the prediction errors for the parameters of the Lotka-Volterra equation.}\label{tab4}%
\begin{tabular}{@{}ccccc@{}}
\toprule
Points & $\alpha$  & $\beta$ & $\delta$ &  $\gamma$  \\
\midrule
\makecell{5}     & \makecell{$2.47 \times 10^{-12}$ \\ $(1.30 \times 10^{-11})$} & \makecell{$2.88 \times 10^{-12}$ \\ $(9.94 \times 10^{-12})$} & \makecell{$7.93 \times 10^{-13}$ \\ $(2.92 \times 10^{-12})$} & \makecell{$1.89 \times 10^{-12}$ \\ $(5.38 \times 10^{-12})$}  \\     
\midrule
\makecell{8}     & \makecell{$4.77 \times 10^{-13}$ \\ $(2.65 \times 10^{-12})$} & \makecell{$5.32 \times 10^{-13}$ \\ $(2.17 \times 10^{-12})$} & \makecell{$1.36 \times 10^{-13}$ \\ $(3.27 \times 10^{-13})$} & \makecell{$3.52 \times 10^{-13}$ \\ $(7.20 \times 10^{-13})$}  \\     
\midrule
\makecell{10}     & \makecell{$4.37 \times 10^{-13}$ \\ $(2.31 \times 10^{-12})$} & \makecell{$4.77 \times 10^{-13}$ \\ $(2.15 \times 10^{-12})$} & \makecell{$1.57 \times 10^{-13}$ \\ $(3.57 \times 10^{-13})$} & \makecell{$3.03 \times 10^{-13}$ \\ $(5.53 \times 10^{-13})$}  \\     
\botrule
\end{tabular}
\end{table}

\begin{table}[htbp]
\caption{Summarize the prediction errors for the parameters of the Lorenz equation.}\label{tab5}%
\begin{tabular}{@{}cccc@{}}
\toprule
Points & $\sigma$  & $\rho$ & $\beta$  \\
\midrule
\makecell{5}     & \makecell{$1.44 \times 10^{-14}$ \\ $(2.80 \times 10^{-13})$} & \makecell{$7.19 \times 10^{-14}$ \\ $(1.21 \times 10^{-13})$} & \makecell{$7.66 \times 10^{-14}$ \\ $(6.82 \times 10^{-14})$}  \\     
\midrule
\makecell{8}     & \makecell{$1.26 \times 10^{-13}$ \\ $(1.63 \times 10^{-13})$} & \makecell{$5.01 \times 10^{-14}$ \\ $(4.68 \times 10^{-14})$} & \makecell{$3.68 \times 10^{-14}$ \\ $(2.74 \times 10^{-14})$}  \\     
\midrule
\makecell{10}     & \makecell{$8.55 \times 10^{-14}$ \\ $(2.0 \times 10^{-13})$} & \makecell{$3.34 \times 10^{-14}$ \\ $(3.28 \times 10^{-14})$} & \makecell{$2.38 \times 10^{-14}$ \\ $(1.84 \times 10^{-14})$}  \\     
\botrule
\end{tabular}
\end{table}

\begin{table}[htbp]
\caption{Summarize the prediction errors for the parameters of the FitzHugh-Nagumo equation.}\label{tab6}%
\begin{tabular}{@{}ccc@{}}
\toprule
Points & $\theta_0$  & $\theta_1$ \\
\midrule
\makecell{5}     & \makecell{$4.80 \times 10^{-05}$ \\ ($1.41 \times 10^{-04}$)} & \makecell{$7.72 \times 10^{-05}$ \\ ($2.25 \times 10^{-04}$)} \\     
\midrule
\makecell{8}     & \makecell{$1.23 \times 10^{-05}$ \\ ($3.27 \times 10^{-05}$)} & \makecell{$3.37 \times 10^{-05}$ \\ ($4.44 \times 10^{-05}$)} \\     
\midrule
\makecell{10}     & \makecell{$1.18 \times 10^{-05}$ \\ ($3.25 \times 10^{-05}$)} & \makecell{$2.08 \times 10^{-05}$ \\ ($4.15 \times 10^{-05}$)} \\     
\botrule
\end{tabular}
\end{table}

\subsection{Compare to RLLPSO}\label{sub4sec3}

For the RLLPSO, the settings of all parameters are consistent with the method proposed in this paper. Table \ref{tab7} presents the mean of MSE across 100 sets of predicted and true parameters by both methods. The values in parentheses represent the SD of the MSE across these 100 sets. 

\begin{table}[htbp]
\centering
\caption{Error results between DERLPSO and RLLPSO.}\label{tab7}
\begin{tabular}{@{}cccc@{}}
\toprule
ODEs & Points & DERLPSO & RLLPSO \\ 
\midrule
\multirow{6}{*}{\vspace*{0.5em}FitzHugh-Nagumo\vspace*{0.5em}}
              & 5  & \makecell{\boldmath{$6.26 \times 10^{-05}$} \\ \boldmath{$(1.8 \times 10^{-04})$}} 
                 & \makecell{0.190 \\ 1.896}  \\
              & 8  & \makecell{\boldmath{$2.3 \times 10^{-05}$} \\ \boldmath{$(3.62 \times 10^{-05})$}} 
                 & \makecell{0.018 \\ (0.177)} \\
              & 10 & \makecell{\boldmath{$1.63 \times 10^{-05}$} \\ \boldmath{$(3.57 \times 10^{-05})$}}  
                 & \makecell{0.013 \\ (0.129)} \\
\midrule
\multirow{6}{*}{\vspace*{0.5em}Lotka-Volterra\vspace*{0.5em}}
              & 5  & \makecell{\boldmath{$2.01 \times 10^{-12}$} \\ \boldmath{$(5.99 \times 10^{-12})$}}  
                 & \makecell{82.26 \\ (310.64)}  \\
              & 8  & \makecell{\boldmath{$3.74 \times 10^{-13}$} \\ \boldmath{$(1.19 \times 10^{-12})$}}  
                 & \makecell{65.51 \\ (138.01)}  \\
              & 10 & \makecell{\boldmath{$3.44 \times 10^{-13}$} \\ \boldmath{$(1.12 \times 10^{-12})$}}  
                 & \makecell{105 \\ (333.93)}  \\
\midrule
\multirow{6}{*}{\vspace*{0.5em}Lorenz\vspace*{0.5em}}
              & 5  & \makecell{$9.75 \times 10^{-14}$ \\ $(1.32 \times 10^{-13})$} 
                 & \makecell{$9.76 \times 10^{-14}$ \\ $(1.32 \times 10^{-13})$}  \\
              & 8  & \makecell{$7.08 \times 10^{-14}$ \\ $(6.27 \times 10^{-14})$}
                 & \makecell{$7.08 \times 10^{-14}$ \\ $(3.26 \times 10^{-14})$}  \\
              & 10 & \makecell{$4.76 \times 10^{-14}$ \\ $(4.77 \times 10^{-14})$}  
                 & \makecell{$4.76 \times 10^{-14}$ \\ $(4.77 \times 10^{-14})$}  \\
\botrule
\end{tabular}
\end{table}

According to Table \ref{tab7}, it can be seen that under different data lengths, for the FitzHugh-Nagumo equation and the Lotka-Volterra equation, the error of DERLPSO is significantly smaller than that of RLLPSO. As the data length increases, the error gradually decreases, indicating that the proposed method exhibits better stability. For the Lorenz equation, the errors of both DERLPSO and RLLPSO are very small and close, which indicates that the proposed method can maintain a high precision comparable to RLLPSO, while also achieving better error control for other equations.

\begin{figure}[htbp]
\centering
\includegraphics[width=0.9\textwidth]{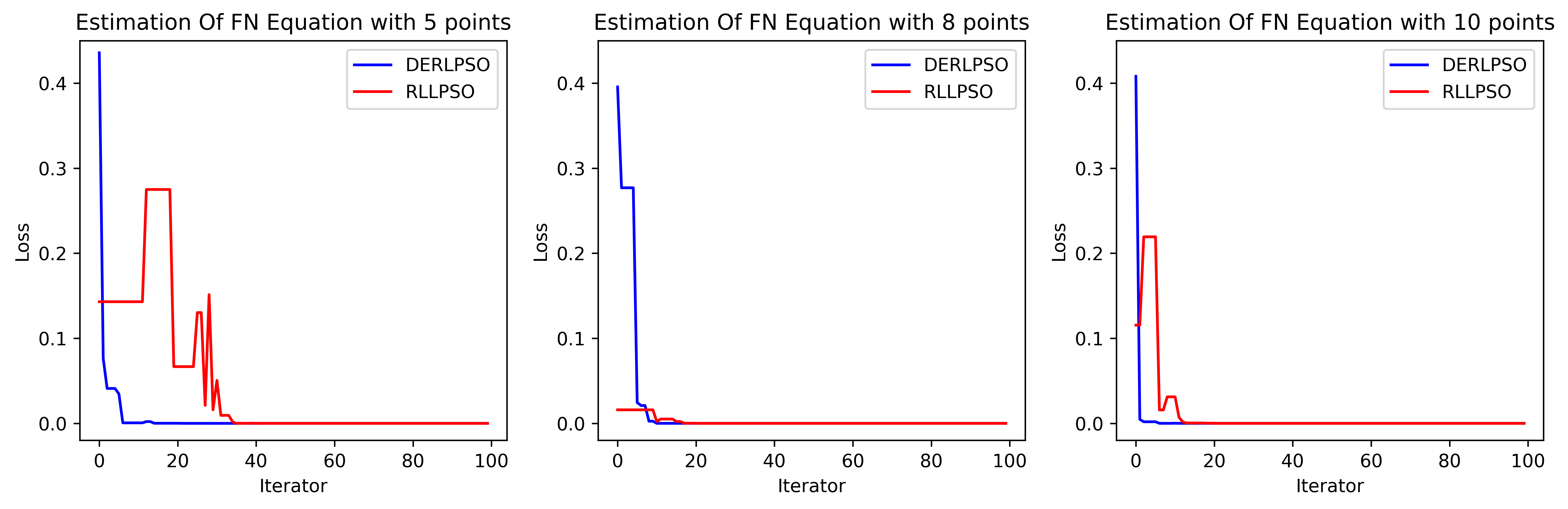}
\caption{Loss curve of ERLPSO and RLLPSO for parameter estimation in the FitzHugh-Nagumo equation.}\label{fig7}
\end{figure}

\begin{figure}[htbp]
\centering
\includegraphics[width=0.9\textwidth]{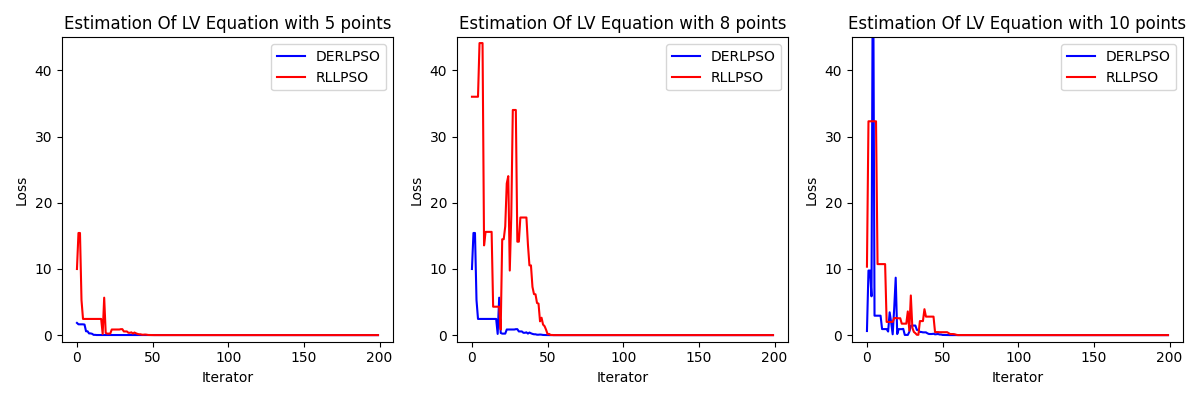}
\caption{Loss curve of ERLPSO and RLLPSO for parameter estimation in the Lotka-Volterra equation.}\label{fig8}
\end{figure}

Figs. \ref{fig6}, \ref{fig7}, and \ref{fig8} show the loss curves for successfully estimating the unknown parameters of three ODEs using the DERLPSO and RLLPSO. The loss value represents the MSE between the true parameters and the predicted parameters. Due to its internal mechanism and optimization strategy, the DERLPSO rapidly reduces the error within fewer iterations, approaching the global optimal solution.

In summary, compared to RLLPSO, DERLPSO can provide more accurate and stable prediction results, with better versatility, and can obtain the global optimal solution through fewer iterations.

\begin{figure}[htbp]
\centering
\includegraphics[width=0.9\textwidth]{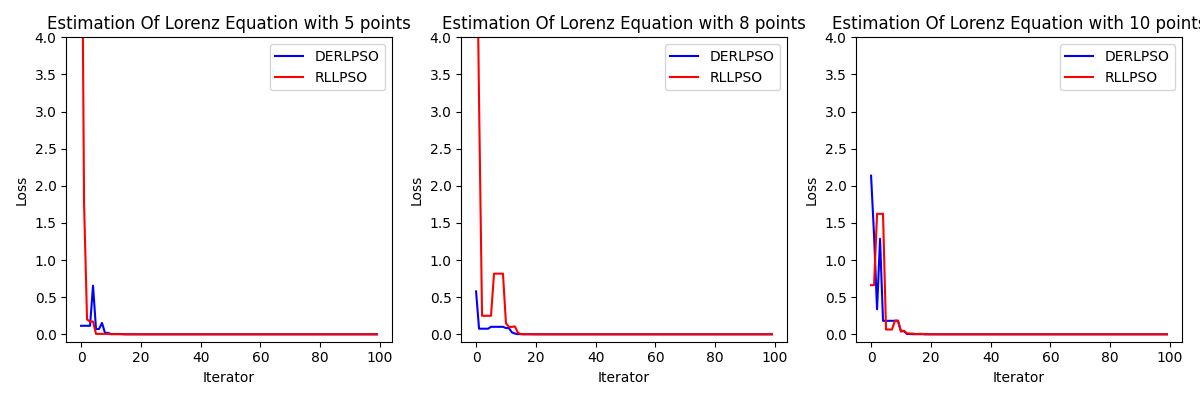}
\caption{Loss curve of ERLPSO and RLLPSO for parameter estimation in the Lorenz equation.}\label{fig9}
\end{figure}

\subsection{Compare to Numerical method}\label{sub5sec3}

The Powell method in numerical methods was proposed by Powell in 1964. It is a search method based on the property that conjugate directions can accelerate convergence speed. This method does not require taking the derivative of the objective function, and can be applied even if the derivative of the objective function is not continuous. Therefore, the Powell method is a highly efficient direct search method.

In DERLPSO, two particle initialization strategies are employed. Thus, when performing calculations using the Powell method, validation was conducted separately for both initialization strategies. The results are shown in Table \ref{tab8}, which presents the mean of MSE across 100 sets of predicted and true parameters. 

Due to the high sensitivity of numerical methods to initial parameter values, it can be observed from Table \ref{tab8} that when the initial parameter values are set randomly, the FitzHugh-Nagumo equations with a logarithmic initialization  strategy achieves a relatively smaller error. However, for the other two ODEs, regardless of the initialization strategy used, the errors obtained are relatively large . In contrast, DERLPSO is able to achieve more accurate and stable prediction results than the Powell method  for multiple equations and different data lengths. The method proposed in this paper not only achieves higher accuracy but also does not rely on the selection of initial parameter values.

\begin{table}[htbp]
\centering
\caption{Error results between DERLPSO and Numerical method.}\label{tab8}
\begin{tabular}{@{}ccccc@{}}
\toprule
\centering ODEs & Points & DERLPSO & RLLPSO(Uniform) & RLLPSO(Logarithmic) \\ 
\midrule
\multirow{3}{*}{\centering FitzHugh-Nagumo}
              & 5  & \makecell{\boldmath{$6.26 \times 10^{-05}$}} 
                 & \makecell{$4.22 \times 10^{18}$} 
                 & \makecell{2.459} \\
              & 8  & \makecell{\boldmath{$2.3 \times 10^{-05}$}} 
                 & \makecell{$8.35 \times 10^{16}$} 
                 & \makecell{0.752} \\
              & 10 & \makecell{\boldmath{$1.63 \times 10^{-05}$}}  
                 & \makecell{$7.20 \times 10^{18}$}
                 & \makecell{0.708} \\
\midrule
\multirow{3}{*}{\centering Lotka-Volterra}
              & 5  & \makecell{\boldmath{$2.01 \times 10^{-12}$}}  
                 & \makecell{$4.42 \times 10^{20}$}
                 & \makecell{$3.71 \times 10^{29}$} \\
              & 8  & \makecell{\boldmath{$3.74 \times 10^{-13}$}}  
                 & \makecell{$1.91 \times 10^{18}$}
                 & \makecell{$2.75 \times 10^{27}$} \\
              & 10 & \makecell{\boldmath{$3.44 \times 10^{-13}$}}  
                 & \makecell{$6.41 \times 10^{15}$}
                 & \makecell{$2.94 \times 10^{29}$} \\
\midrule
\multirow{3}{*}{\centering Lorenz}
              & 5  & \makecell{\boldmath{$9.75 \times 10^{-14}$}} 
                 & \makecell{$3.74 \times 10^{12}$}
                 & \makecell{$2.09 \times 10^{11}$} \\
              & 8  & \makecell{\boldmath{$7.08 \times 10^{-14}$}}
                 & \makecell{$4.94 \times 10^{11}$}
                 & \makecell{$4.96 \times 10^{12}$} \\
              & 10 & \makecell{\boldmath{$4.76 \times 10^{-14}$}}  
                 & \makecell{$1.01 \times 10^{13}$}
                 & \makecell{$2.81 \times 10^{12}$} \\
\botrule
\end{tabular}
\end{table}

\subsection{Compare to Deep learning}\label{sub6sec3}

Neural ODEs \cite{bib15} can effectively solve the ODEs present in data. Since this model cannot directly extract the equation parameters, this paper modifies Neural ODEs to enable the prediction of unknown parameters in ODEs. 
In deep learning methods, Fully Connected Neural Networks (FCNNs), possess strong abilities in fitting multivariate and high-dimensional functions. Among all available neural networks, Recurrent Neural Networks (RNNs)stand out in modeling time series data problems due to their gated units and memory storage capabilities \cite{bib25}. Encoder-Decoder or Variational Autoencoder (VAEs) models also have the ability to extract latent features from data. Therefore, this paper also uses classical deep learning methods such as FCNNs, RNNs, and VAEs to solve the parameters of ODEs. Notably, the Encoder and Decoder of the VAEs are both composed of Neural ODEs, with the network structure presented in Appendix A.

Tables \ref{tab9} and \ref{tab10} present the mean of MSE across 100 sets of predicted and true parameters by DERLPSO and four deep learning methods. The values in parentheses represent the SD of the MSE across these 100 sets.\

Deep learning methods rely on training data to learn. The experiments first uses training and testing data generated from the same Gaussian distribution for parameter sampling. The related parameters are shown in Table \ref{tab1}, and the experimental results are presented in Table \ref{tab9}. However, during the training process, neural networks tend to learn specific patterns inherent in the training data, which hinders the effective evaluation of the generalization ability of deep learning methods. 
To circumvent this issue, the experiments also employs a different strategy, where the training data is generated by sampling parameters from a Gaussian distribution with a variance of 0.1, while the testing data is generated by sampling parameters from a Gaussian distribution with a variance of 0.5. This method allows for a more comprehensive evaluation of the models, and the experimental results are shown in Table \ref{tab10}.

\begin{table}[htbp]
\centering
\caption{Error results between DERLPSO and Deep Learning Methods under the same parameter distribution.}\label{tab9}
\begin{tabular}{@{}ccccccc@{}}
\toprule
ODEs & Points & DERLPSO & FCNNs	 &  RNNs  &  Neural ODEs  &  VAEs \\ 
\midrule
\multirow{6}{*}{\vspace*{0.5em}FitzHugh-Nagumo\vspace*{0.5em}}
              & 5  & \makecell{\boldmath{$6.26 \times 10^{-05}$} \\ \boldmath{$(1.8 \times 10^{-04})$}} 
                 & \makecell{0.014 \\ (0.020)}
                 & \makecell{0.050 \\ (0.073)}
                 & \makecell{0.029 \\ (0.042)}
                 & \makecell{0.038 \\ (0.040)} \\
              & 8  & \makecell{\boldmath{$2.3 \times 10^{-05}$} \\ \boldmath{$(3.62 \times 10^{-05})$}}
                 & \makecell{0.017 \\ (0.062)}
                 & \makecell{0.037 \\ (0.084)}
                 & \makecell{0.021 \\ (0.063)}
                 & \makecell{0.099 \\ (0.124)} \\
              & 10 & \makecell{\boldmath{$1.63 \times 10^{-05}$} \\ \boldmath{$(3.57 \times 10^{-05})$}}
                 & \makecell{0.018 \\ (0.067)}
                 & \makecell{0.037 \\ (0.079)}
                 & \makecell{0.021 \\ (0.067)}
                 & \makecell{0.085 \\ (0.13)} \\
\midrule
\multirow{6}{*}{\vspace*{0.5em}Lotka-Volterra\vspace*{0.5em}}
              & 5  & \makecell{\boldmath{$2.01 \times 10^{-12}$} \\ \boldmath{$(5.99 \times 10^{-12})$}}  
                 & \makecell{0.050 \\ (0.073)}
                 & \makecell{0.151 \\ (0.127)}
                 & \makecell{0.143 \\ (0.127)}
                 & \makecell{0.103 \\ (0.107)}  \\
              & 8  & \makecell{\boldmath{$3.74 \times 10^{-13}$} \\ \boldmath{$(1.19 \times 10^{-12})$}}
                 & \makecell{0.063 \\ (0.075)}
                 & \makecell{0.165 \\ (0.130)}
                 & \makecell{0.186 \\ (0.139)}
                 & \makecell{0.096 \\ (0.090)}  \\
              & 10 & \makecell{\boldmath{$3.44 \times 10^{-13}$} \\ \boldmath{$(1.12 \times 10^{-12})$}}  
                 & \makecell{0.067 \\ (0.071)}
                 & \makecell{0.149 \\ (0.113)}
                 & \makecell{0.163 \\ (0.120)}
                 & \makecell{0.122 \\ (0.097)}  \\
\midrule
\multirow{6}{*}{\vspace*{0.5em}Lorenz\vspace*{0.5em}}
              & 5  & \makecell{\boldmath{$9.75 \times 10^{-14}$} \\ \boldmath{$(1.32 \times 10^{-13})$}}
                 & \makecell{0.093 \\ (0.129)}
                 & \makecell{0.114 \\ (0.132)}
                 & \makecell{0.081 \\ (0.114)}
                 & \makecell{0.067 \\ (0.130)} \\
              & 8  & \makecell{\boldmath{$7.08 \times 10^{-14}$} \\ \boldmath{$(6.27 \times 10^{-14})$}}
                 & \makecell{0.015 \\ (0.033)}
                 & \makecell{0.161 \\ (0.161)}
                 & \makecell{0.099 \\ (0.118)}
                 & \makecell{0.098 \\ (0.175)} \\
              & 10 & \makecell{\boldmath{$4.76 \times 10^{-14}$} \\ \boldmath{$(4.77 \times 10^{-14})$}}
                 & \makecell{0.079 \\ (0.121)}
                 & \makecell{0.165 \\ (0.185)}
                 & \makecell{0.173 \\ (0.210)}
                 & \makecell{0.077 \\ (0.130)} \\
\botrule
\end{tabular}
\end{table}

\begin{table}[htbp]
\centering
\caption{Error results between DERLPSO and Deep Learning Methods under different parameter distributions.}\label{tab10}
\begin{tabular}{@{}ccccccc@{}}
\toprule
ODEs & Points & DERLPSO & FCNNs & RNNs & Neural ODEs & VAEs \\ 
\midrule
\multirow{6}{*}{\vspace*{0.5em}FitzHugh-Nagumo\vspace*{0.5em}}
              & 5  & \makecell{\boldmath{$6.26 \times 10^{-05}$} \\ \boldmath{$(1.8 \times 10^{-04})$}} 
                 & \makecell{0.513 \\ (0.618)}
                 & \makecell{0.319 \\ (0.633)}
                 & \makecell{0.129 \\ (0.164)}
                 & \makecell{0.152 \\ (0.177)} \\
              & 8  & \makecell{\boldmath{$2.3 \times 10^{-05}$} \\ \boldmath{$(3.62 \times 10^{-05})$}}
                 & \makecell{0.203 \\ (0.430)}
                 & \makecell{0.382 \\ (0.685)}
                 & \makecell{0.182 \\ (0.239)}
                 & \makecell{0.219 \\ (0.229)} \\
              & 10 & \makecell{\boldmath{$1.63 \times 10^{-05}$} \\ \boldmath{$(3.57 \times 10^{-05})$}}
                 & \makecell{0.440 \\ (0.900)}
                 & \makecell{0.258 \\ (0.464)}
                 & \makecell{0.182 \\ (0.293)}
                 & \makecell{0.248 \\ (0.336)} \\
\midrule
\multirow{6}{*}{\vspace*{0.5em}Lotka-Volterra\vspace*{0.5em}}
              & 5  & \makecell{\boldmath{$2.01 \times 10^{-12}$} \\ \boldmath{$(5.99 \times 10^{-12})$}}  
                 & \makecell{0.185 \\ (0.313)}
                 & \makecell{0.214 \\ (0.165)}
                 & \makecell{0.216 \\ (0.182)}
                 & \makecell{0.155 \\ (0.157)} \\
              & 8  & \makecell{\boldmath{$3.74 \times 10^{-13}$} \\ \boldmath{$(1.19 \times 10^{-12})$}}
                 & \makecell{0.218 \\ (0.490)}
                 & \makecell{0.191 \\ (0.164)}
                 & \makecell{0.159 \\ (0.148)}
                 & \makecell{0.158 \\ (0.173)} \\
              & 10 & \makecell{\boldmath{$3.44 \times 10^{-13}$} \\ \boldmath{$(1.12 \times 10^{-12})$}}  
                 & \makecell{0.173 \\ (0.171)}
                 & \makecell{0.226 \\ (0.145)}
                 & \makecell{0.208 \\ (0.137)}
                 & \makecell{0.191 \\ (0.173)} \\
\midrule
\multirow{6}{*}{\vspace*{0.5em}Lorenz\vspace*{0.5em}}
              & 5  & \makecell{\boldmath{$9.75 \times 10^{-14}$} \\ \boldmath{$(1.32 \times 10^{-13})$}}
                 & \makecell{0.119 \\ (0.177)}
                 & \makecell{0.167 \\ (0.151)}
                 & \makecell{0.172 \\ (0.144)}
                 & \makecell{0.208 \\ (0.167)} \\
              & 8  & \makecell{\boldmath{$7.08 \times 10^{-14}$} \\ \boldmath{$(6.27 \times 10^{-14})$}}
                 & \makecell{0.132 \\ (0.178)}
                 & \makecell{0.152 \\ (0.162)}
                 & \makecell{0.419 \\ (0.267)}
                 & \makecell{0.196 \\ (0.172)} \\
              & 10 & \makecell{\boldmath{$4.76 \times 10^{-14}$} \\ \boldmath{$(4.77 \times 10^{-14})$}}
                 & \makecell{0.082 \\ (0.158)}
                 & \makecell{0.189 \\ (0.199)}
                 & \makecell{0.206 \\ (0.184)}
                 & \makecell{0.155 \\ (0.160)} \\
\botrule
\end{tabular}
\end{table}

According to Tables \ref{tab9} and \ref{tab10}, it can be observed that while training and predicting with the same parameter distribution yields smaller errors, DERLPSO consistently achieves lower errors compared to deep learning methods in both cases, across various equations and data lengths. Furthermore, it has two notable advantages. Firstly, it does not require prior model training, making it easily applicable to other equations. Secondly, it does not require large amounts of data, whereas deep learning methods typically require extensive training data. DERLPSO is not limited by training data and can handle various unknown data distributions with greater flexibility, easily adapting to other ODEs and demonstrating exceptional versatility.

\subsection{Compare to Bayesian method}\label{sub7sec3}

This paper compares the performance between DEPLPSO and a Bayesian method based on MCMC sampling \cite{bib11} on the Lotka-Volterra equations. The results are shown in Table \ref{tab11}, which presents the mean of MSE across 100 sets of predicted and true parameters using both DERLPSO and the Bayesian method. The values in parentheses represent the SD of the MSE across these 100 sets.

\begin{table}[htbp]
\centering
\caption{Error results between DERLPSO and Bayesian Method.}\label{tab11}
\begin{tabular}{@{}cccc@{}}
\toprule
ODEs & Points & DERLPSO & Bayesian \\ 
\midrule
\multirow{6}{*}{\vspace*{0.5em}Lotka-Volterra\vspace*{0.5em}}
              & 5  & \makecell{\boldmath{$2.01 \times 10^{-12}$} \\ \boldmath{$(5.99 \times 10^{-12})$}} 
                 & \makecell{0.136 \\ (0.193)} \\
              & 8  & \makecell{\boldmath{$3.74e \times 10^{-13}$} \\ \boldmath{$(1.19 \times 10^{-12})$}}
                 & \makecell{0.402 \\ (1.711)} \\
              & 10 & \makecell{\boldmath{$3.44 \times 10^{-13}$} \\ \boldmath{$(1.12 \times 10^{-12})$}}
                 & \makecell{0.273 \\ (2.606)} \\
\botrule
\end{tabular}
\end{table}

The Bayesian method achieves the smallest prediction error when the time series data length is five points, but it is still higher than the DERLPSO method proposed in this paper. Furthermore, the Bayesian method replaces the ODEs constraints with probabilistic expressions and combines them with a non-parametric data fitting process into a joint likelihood framework, MCMC sampling is then used to sample from the joint posterior distribution, requiring the definition of a probabilistic expression specific to the equation structure. DEPLPSO possesses stronger versatility while achieving good accuracy compared to the Bayesian method.

\subsection{DERLPSO for PDEs examples}\label{sub8sec3}

\begin{figure}[htbp]
\centering
\includegraphics[width=0.9\textwidth]{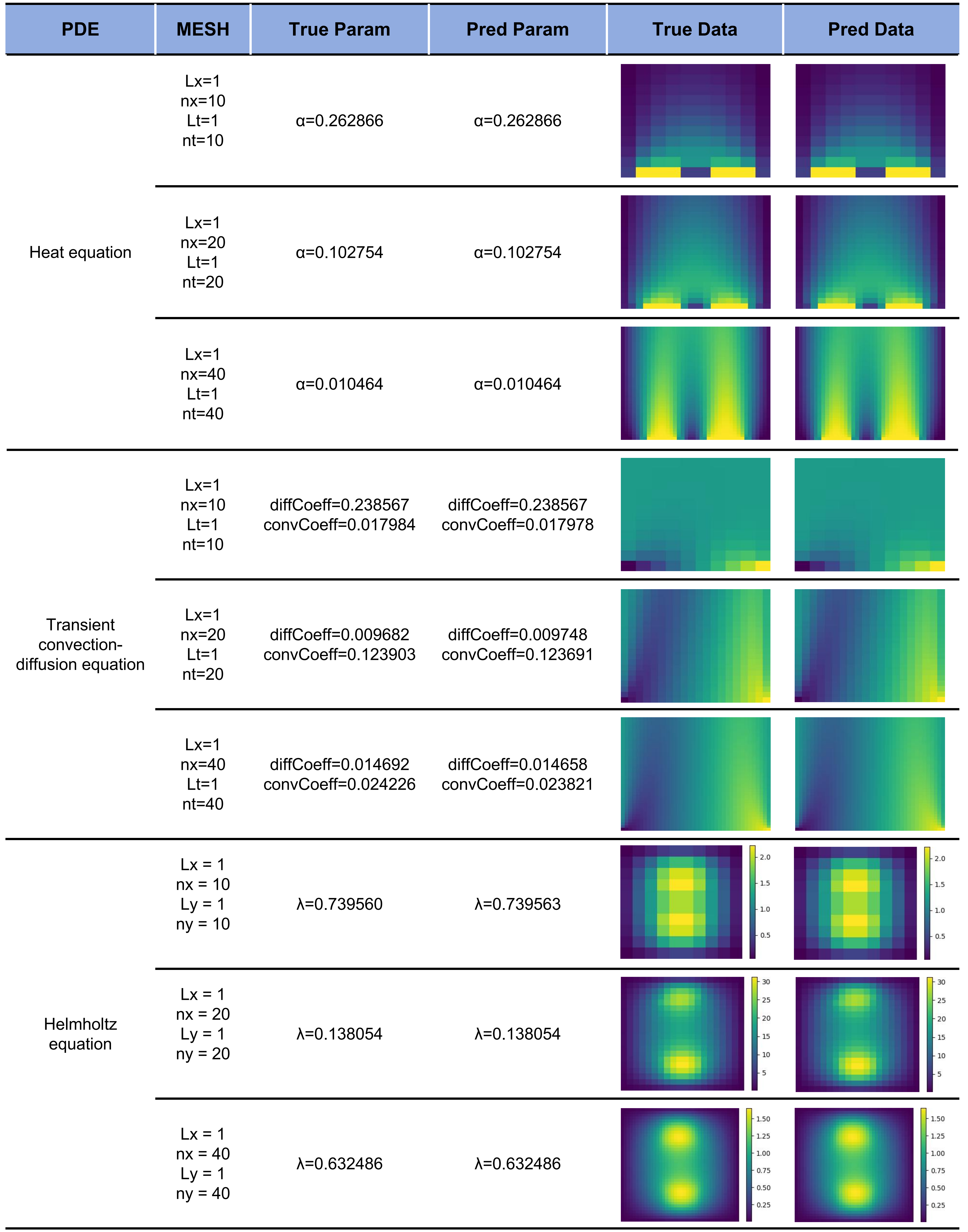}
\caption{Comparison of True and Predicted Data in PDEs Examples and Corresponding Experimental Scenarios.}\label{fig10}
\end{figure}

Table \ref{tab12} shows the mean of MSE across 100 sets of predicted and true
parameters for the Heat equation, Transient Convection-Diffusion equation, and Helmholtz equation. The values in parentheses represent the SD of the MSE across these 100 sets. It can be observed that DEPLPSO provides reasonable predictions for the unknown parameters of the above three types of PDEs, with small MSE and SD values, indicating that the method has good accuracy and stability, and can well solve the unknown parameters of PDEs. Fig. \ref{fig10} shows the comparison between the simulated data and the fitted data under different PDEs and experimental scenarios.

\begin{table}[htbp]
\centering
\caption{Summary of prediction errors for parameters in three PDEs.}\label{tab12}
\begin{tabular}{@{}cccccc@{}}
\toprule
Points & Heat & \multicolumn{2}{c}{Transient Convection-Diffusion} & Helmholtz \\ 
\midrule
 & $\alpha$ & D & v & $\lambda$ \\ 
\midrule
5  & \makecell{$7.4 \times 10^{-09}$ \\ $(1.0 \times 10^{-08})$}
   & \makecell{$2.8 \times 10^{-04}$ \\ $(5.2 \times 10^{-04})$}
   & \makecell{$4.2 \times 10^{-05}$ \\ $(8.6 \times 10^{-05})$} 
   & \makecell{$1.49 \times 10^{-05}$ \\ $(1.9 \times 10^{-05})$} \\
\midrule
8  & \makecell{$1.2 \times 10^{-08}$ \\ $(1.5 \times 10^{-08})$} 
   & \makecell{$2.0 \times 10^{-04}$ \\ $(4.0 \times 10^{-04})$}
   & \makecell{$1.6 \times 10^{-04}$ \\ $(4.2 \times 10^{-04})$} 
   & \makecell{$9.63 \times 10^{-06}$ \\ $(1.19 \times 10^{-05})$} \\
\midrule
10 & \makecell{$1.3 \times 10^{-08}$ \\ $(1.6 \times 10^{-08})$} 
   & \makecell{$1.9 \times 10^{-04}$ \\ $(2.9 \times 10^{-04})$}
   & \makecell{$3.3 \times 10^{-05}$ \\ $(5.1 \times 10^{-05})$} 
   & \makecell{$3.43 \times 10^{-04}$ \\ $(9.6 \times 10^{-04})$} \\
\botrule
\end{tabular}
\end{table}

\section{Conclusions}\label{sec4}

This paper has improved RLLPSO and proposed the DERLPSO method for solving the unknown parameters of differential equations. This method can obtain a lower error compared to the RLLPSO. Additionally, DERLPSO can avoid the disadvantages of traditional numerical methods, such as being sensitive to initial parameter values and easily getting trapped in local optima. Compared to deep learning methods for solving parameters in differential equations, not only can it achieve higher accuracy, but it also does not require pre-training with large amounts of data. Compared to Bayesian method, it can avoid the need to unfer the specific equation structure. In summary, the DERLPSO method proposed in this paper performs well in solving unknown parameters of differential equations, with advantages such as high accuracy, strong versatility, and independence from initial parameter values.

It is well-known that as the complexity of the solution space of differential equations increases with the number of variables. Therefore, one direction for future work is to improve the algorithm so that it can accurately infer the unknown parameters of more complex and larger-scale differential equations.

Although the method proposed in this paper demonstrates excellent global search capability and robustness in solving the unknown parameter problem of differential equations, it does not show significant improvement in computational speed compared to traditional methods. This issue limits the practical application potential of the method in high-dimensional complex models that requiring rapid iterations. Future research could consider incorporating parallel computing, distributed optimization techniques, or exploring hybrid optimization algorithms to address the current bottleneck in computational speed.

Stochastic differential equations (SDEs) have become the standard model for diffusive processes in physics and biological sciences, as well as economics and ﬁnance \cite{bib26}. The randomness in SDEs poses challenges for the accuracy of parameter estimation. In this paper, particles can optimize their direction based on feedback from reinforcement learning during each update, thereby better handling random fluctuations and improving the stability and precision of the estimates. Traditional methods for parameter estimation in SDEs may require extensive computation to obtain desirable results. The method proposed in this paper enables real-time optimization of particle trajectories, thereby accelerating the convergence process of PSO. Therefore, this method provides a novel and efficient solution for parameter estimation in SDEs, with potential applicability in dealing with nonlinear and high-dimensional SDEs.

\bibliography{sn-bibliography}

\newpage

\begin{appendices}

\section{Extra Figures}\label{secA1}

This appendix provides model architecture diagrams for the four methods used in this paper: FCNNs, RNNs, VAEs, and Neural ODEs.

\begin{figure}[htbp]
\centering
\includegraphics[width=1\textwidth]{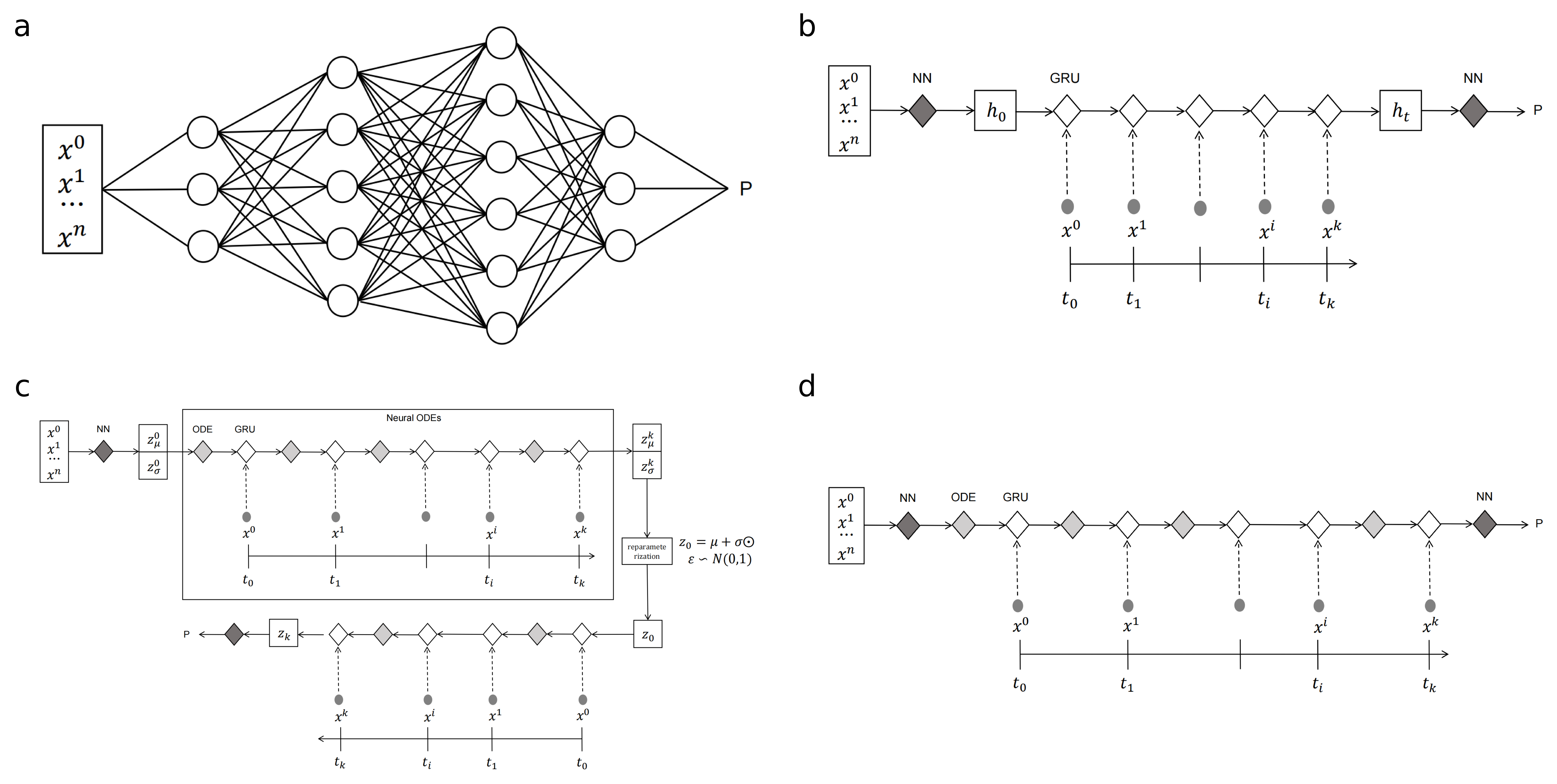}
\caption{Network structure diagrams used in this paper. (a).FCNNs structure diagram used in this paper: A fully connected feedforward neural network comprises an input layer, hidden layers, and an output layer. The input layer receives input features $x_0, x_1, \ldots, x_n$, which are transmitted through fully connected pathways to the hidden layers. The network consists of multiple hidden layers, with each node in a hidden layer applying a nonlinear transformation to the inputs using activation functions (such as ReLU or Sigmoid). Each node in the hidden layers is fully connected to all nodes in both the preceding and subsequent layers, forming a complete fully connected structure. Ultimately, the output layer generates the network's prediction result $P$. (b).RNNs structure diagram used in this paper: A combined Gated Recurrent Unit (GRU) Time series neural network structure. The input features $x_0, x_1, \ldots, x_n$ are processed through a feedforward neural network (NN) to obtain the initial hidden state $h_0$. Subsequently, the GRU layer processes the time series data, progressively updating the hidden state $h_t$ through a gating mechanism to capture the temporal dependencies of the input data. At each time step $t_0, t_1, \ldots, t_k$, the GRU updates the hidden state, and finally, another feedforward neural network maps the last hidden state $h_t$ to the predicted output $P$. (c).VAEs structure diagram used in this paper: Both the Encoder and Decoder are specified as Neural ODEs. In the VAEs, the Encoder encodes the time series data to obtain the distribution of the latent space, followed by reparameterization to sample and generate the initial state input $z_0$ for the Decoder. The dimensionality of $z_0$ may not align with the parameters to be predicted. After the Neural ODEs process all the time series data, the final output state value $z_k$ is fed into a fully connected network to obtain the parameters $P$ that need to be predicted. (d).Neural ODEs structure diagram used in this paper: A state generation layer and a parameter extraction layer are added before and after the Neural ODEs, respectively. The GRU is utilized, and during the ODE process, the ODE Solver employs the dopri5 method from the SciPy library in the experiments.}\label{figA1}
\end{figure}

\end{appendices}

\end{document}